\documentclass[runningheads]{llncs}

\usepackage{accv}

\usepackage{accvabbrv}

\usepackage{graphicx}
\usepackage{booktabs}

\usepackage[accsupp]{axessibility}  % Improves PDF readability for those with disabilities.

\usepackage{hyperref}

\usepackage{orcidlink}

\usepackage{graphicx}
\usepackage{amsmath,amssymb} % define this before the line numbering.
\usepackage{color}
\usepackage{booktabs}

\usepackage{xcolor} 
\usepackage{booktabs} % For professional looking tables

\usepackage[accsupp]{axessibility}  % Improves PDF readability for those with disabilities.

\usepackage{epsfig}
\usepackage[normalem]{ulem}
\usepackage{sidecap}
\usepackage{tabularx} % Automatically calculates column widths with text wrapping
\usepackage{algorithm,algorithmicx,algpseudocode}
\usepackage{adjustbox}
\usepackage{multirow}
\usepackage{comment}
\def\etal{\textit{et al.}}

\begin{document}

% ---------------------------------------------------------------
% TODO REVIEW: Replace with your title
\title{Foundation Model-Powered 3D Few-Shot Class Incremental Learning via Training-free Adaptor} 

% TODO REVIEW: If the paper title is too long for the running head, you can set
% an abbreviated paper title here. If not, comment out.
\titlerunning{Training-Free Foundation Model for 3D FSCIL}

% TODO FINAL: Replace with your author list. 
% Include the authors' OCRID for the camera-ready version, if at all possible.
\author{Sahar Ahmadi \inst{1}\orcidlink{0000-0002-2161-7005} \and
Ali Cheraghian\inst{2,3}\orcidlink{0000-0002-3324-7849} \and
Morteza Saberi\inst{1}\orcidlink{0000-0002-5168-2078} \and 
Md.Towsif Abir\inst{4}\orcidlink{0000-0003-1198-5098} \and
Hamidreza Dastmalchi \inst{5}\orcidlink{0000-0003-2818-4182} \and
Farookh Hussain\inst{1}\orcidlink{0000-0003-1513-8072} \and
Shafin Rahman\inst{4}\orcidlink{0000-0001-7169-0318}}

% TODO FINAL: Replace with an abbreviated list of authors.
%\authorrunning{F.~Author et al.}
% First names are abbreviated in the running head.
% If there are more than two authors, 'et al.' is used.

\authorrunning{S. Ahmadi et al.}
\institute{\textsuperscript{1} University of Technology Sydney, Australia \textsuperscript{2} Data61, CSIRO, Australia \\ \textsuperscript{3} Australian National University, Australia \textsuperscript{4} North South University, Bangladesh \\ \textsuperscript{5} York University, Canada
\\
% Corresponding Author Email: \email{\small\texttt{shafin.rahman@northsouth.edu}}
\email{\small\texttt{\{sahar.ahmadi, morteza.saberi, farookh.hussain\}@uts.edu.au,\\\{shafin.rahman, towsif.abir\}@northsouth.edu, ali.cheraghian@data61.csiro.au, hrd@yorku.ca}}
}

% TODO FINAL: Replace with your institution list.
% \institute{Princeton University, Princeton NJ 08544, USA \and
% Springer Heidelberg, Tiergartenstr.~17, 69121 Heidelberg, Germany
% \email{lncs@springer.com}\\
% \url{http://www.springer.com/gp/computer-science/lncs} \and
% ABC Institute, Rupert-Karls-University Heidelberg, Heidelberg, Germany\\
% \email{\{abc,lncs\}@uni-heidelberg.de}}

\maketitle

\begin{abstract}
Recent advances in deep learning for processing point clouds hold increased interest in Few-Shot Class Incremental Learning (FSCIL) for 3D computer vision. This paper introduces a new method to tackle the Few-Shot Continual Incremental Learning (FSCIL) problem in 3D point cloud environments. We leverage a foundational 3D model trained extensively on point cloud data. Drawing from recent improvements in foundation models, known for their ability to work well across different tasks, we propose a novel strategy that does not require additional training to adapt to new tasks. Our approach uses a dual cache system: first, it uses previous test samples based on how confident the model was in its predictions to prevent forgetting, and second, it includes a small number of new task samples to prevent overfitting. This dynamic adaptation ensures strong performance across different learning tasks without needing lots of fine-tuning. We tested our approach on datasets like ModelNet, ShapeNet, ScanObjectNN, and CO3D, showing that it outperforms other FSCIL methods and demonstrating its effectiveness and versatility. The code is available at \url{https://github.com/ahmadisahar/ACCV_FCIL3D}.

\end{abstract}

\section{Introduction}

In recent years, point cloud processing based on deep learning models has become a crucial research direction in computer vision due to its wide range of potential applications in real-world scenarios. Despite significant progress in this field \cite{pointnet2017, pointnet++2017, RS-CNN2019, zhao2021point}, much of the research has been carried out in controlled environments. When designing a point cloud classification model, it is practical to consider scenarios where data for all classes cannot be collected simultaneously. Typically, we start with numerous training samples for some classes, termed the base task, to develop a baseline model, and then gradually collect data for the remaining classes, termed novel tasks. Additionally, due to hardware limitations or privacy concerns, retraining the model with base task data may not be feasible when adapting the baseline model to novel task samples. This situation leads to the problem of catastrophic forgetting, where the model tends to forget the old classes while learning the new ones. Furthermore, data collection for new classes is often limited and we may not obtain more than a few samples for new classes, leading to overfitting issues in novel classes. The combination of these two issues is studied in the literature under the umbrella of Few-Shot Class Incremental Learning (FSCIL). Specifically for 3D point cloud data, the base task usually consists of synthetic data. In contrast, the novel tasks consist of real scan data, leading to domain gap issues that add more complexity to the FSCIL problem for 3D point cloud data than the 2D image domain.

\begin{figure*}[!t]
\centering
\includegraphics[width=1\textwidth]{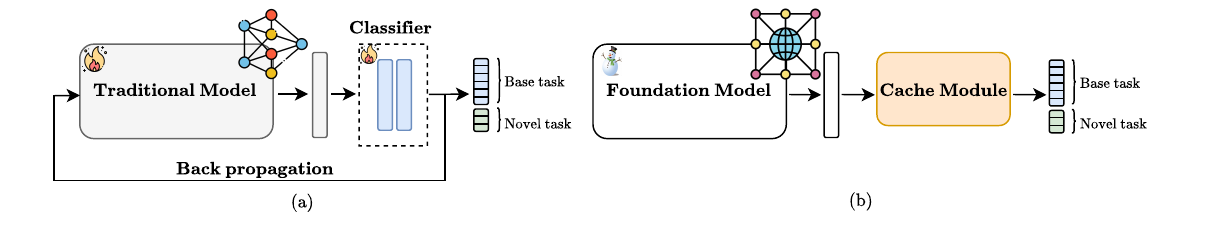}  
\caption{(a) Existing methods~\cite{chen2021incremental, chowdhury2022few, cheraghian2021semantic} for FSCIL typically employ a traditional vision model trained from scratch on the base task, followed by a classifier. Adding novel classes requires fine-tuning with a few novel training samples, often overfitting the novel classes and forgetting the base classes. (b) In contrast, our proposed FSCIL strategy leverages a foundation model pre-trained on a large dataset, which offers strong generalization with minimal effort compared to traditional vision models. Specifically, to incorporate novel classes into the base classes, we introduce a novel strategy that eliminates the need for fine-tuning, thereby reducing both forgetting and overfitting issues. Instead, we use a novel training-free adaptation module to seamlessly integrate novel classes with existing base classes with minimal effort.}
\label{model_motivation}
\end{figure*}

Existing methods~\cite{tao2020few, chowdhury2022few, Tan_2024_WACV} (see Fig.~\ref{model_motivation} (a)) to address the forgetting problem in FSCIL often rely on rehearsal strategies to mitigate the forgetting issue. This approach entails replaying samples from old classes, usually stored in memory, while learning the novel classes to address the forgetting problem. Additionally, to train the few-shot novel classes, they~\cite{cheraghian2021semantic, mazumder2021few, Cheraghian_2021_ICCV} usually fine-tune a base model through back-propagation, previously trained on the base task, to learn the few-shot novel classes. Unfortunately, this strategy leads to the overfitting issue of novel classes. In contrast, this paper (see Fig.~\ref{model_motivation} (b)) introduces a novel training-free adaptation strategy applied on top of a foundational model, eliminating the need to fine-tune the base model and thus avoiding common issues in FSCIL such as forgetting and overfitting. Specifically, our approach ensures that the base model remains intact while efficiently learning new classes without compromising the knowledge of previously learned ones. This preserves the integrity of the base model and enhances its ability to generalize on few-shot novel classes.

In this paper, we leverage a 3D foundation model~\cite{zhou2023uni3d}, trained on an extensive number of 3D point cloud samples, to tackle the FSCIL problem setting. Foundation models in the literature have demonstrated powerful generalization across incremental tasks with minimal effort~\cite{Cao_2024_CVPR, Kim_2024_CVPR}. Moreover, to address the forgetting and overfitting issues in FSCIL, we propose a novel training-free adaptation strategy that eliminates the need to fine-tune the foundation model for novel tasks. Specifically, to mitigate forgetting the base task, we introduce a novel mechanism to leverage test samples from the base task during inference. These test samples are selected on the basis of the confidence score of the foundation model and stored in a cache for later use during inference. Additionally, to control overfitting, we maintain few-shot samples from novel tasks in the adaptor. This dual cache approach ensures the model remains robust against forgetting, while efficiently learning new classes without overfitting.

Overall, the main contributions of our proposed method are:
\begin{itemize}
    \item We leverage the power of a 3D foundation model, applying it for the first time to the 3D FSCIL task. 
    \item We introduce a novel training-free adaptation strategy that utilizes the incoming test samples to dynamically adapt the model for future test samples. This approach helps maintain performance in previously learned classes while effectively learning new ones.
    \item We achieve state-of-the-art results in three cross-data set settings, demonstrating the robustness and generalizability of our method.
\end{itemize}

\section{Related work}

\noindent\textbf{Point cloud processing:} Previous deep learning approaches for 3D point clouds primarily addressed the learning problem by converting the point cloud data into intermediate representations. These methods included rendering the 3D point cloud into 2D images \cite{qi2016volumetric, su2015multi}, or constructing meshes \cite{bruna2013spectral, masci2015geodesic} for further processing. However, these approaches were constrained by their limited ability to accurately represent and understand complex 3D scenes and non-isometric shapes \cite{pointnet2017}.
PointNet \cite{pointnet2017} was a pioneering work that explored the direct processing of 3D point clouds without any intermediate representation. However, by design, PointNet overlooked the local structures induced by the distance metric. PointNet++ \cite{pointnet++2017} effectively resolved this issue by processing sets of points sampled in a metric space in a hierarchical fashion, allowing the network to capture local structures more accurately. Building upon this, several studies~\cite{RS-CNN2019,spidercnn2018,SPHNet2019,SFCNN2019,pointconv2019,pointcnn2018} have suggested convolutional techniques designed to extract local features. PointConv \cite{pointconv2019} introduced an inverse density scale to re-weight the continuous function learned by MLP, which corresponds to the Monte Carlo approximation of the continuous 3D convolution.

Simultaneously, other research efforts~\cite{dgcnn2019,LocalSpecGCN2018,PointGCN2018} considered each point cloud as a graph vertex to extract features in spatial or spectral domains. For example, DGCNN \cite{dgcnn2019} proposed a method that dynamically computes graphs at each layer of the neural network, improving the representation power of point clouds by capturing local geometric structures and recovering topology. In contrast, PointGCN~\cite{PointGCN2018} leveraged localized graph convolutions with two types of graph downsampling operations to effectively explore the local structure of point clouds. To address the challenges posed by the irregular and unordered nature of point cloud data, Guo~\etal{}~\cite{guo2021pct} introduced a framework based on the Transformer architecture, which has achieved success in natural language processing and image processing.

\noindent\textbf{Few-shot class incremental learning:} Tao et al.~\cite{tao2020few} pioneered FSCIL framework for image data. The authors proposed a framework that leveraged a neural gas (NG) network to preserve the topology of the feature manifold formed by different classes, stabilizing the old class knowledge and improving representation learning for few-shot new classes. In a subsequent work~\cite{Cheraghian_2021_ICCV}, a novel approach was proposed that extends inductive zero-shot learning (ZSL) to transductive ZSL and Generalized ZSL (GZSL) for 3D point cloud classification while addressing challenges related to domain adaptation, hubness, and data bias. Cheraghian et al.~\cite{cheraghian2021semantic} proposed a novel vision-language approach, integrating class semantic information from language space using distillation to mitigate catastrophic forgetting, along with an attention mechanism to address overfitting on few-shot novel tasks. FSCIL3D~\cite{chowdhury2022few} introduced the innovative concept of Microshape. By leveraging Microshapes, the model could handle incremental training with few-shot examples more effectively, bridging the gap between synthetic and real data. The work by Tan et al.~\cite{Tan_2024_WACV} explored cross-domain FSCIL applied to point-cloud recognition, where their base model discriminates between base samples (treated as in-distribution) and new samples (considered out-of-distribution).

\noindent\textbf{Foundation models:} Foundation models represent a significant evolution in the field of computer vision, distinguished by their ability to generalize across a wide range of tasks and modalities. These models are trained using vast datasets that span various domains, which imbue them with unprecedented flexibility and capability to handle diverse applications, from image recognition to multimodal reasoning that combines text, images, and audio data. The core architectural innovations in foundation models, such as dual encoders and sophisticated fusion mechanisms, allow efficient integration and processing of multimodal information, enabling these models to perform tasks with a degree of sophistication that mirrors human cognitive abilities\cite{awais2023foundational}. One of the most notable features of foundation models is their proficiency in `zero-shot' learning, where the model applies the knowledge acquired during training to new tasks it has never explicitly learned. For instance, models like CLIP can accurately classify images or generate descriptions based on textual prompts without direct training on those specific tasks. This capability not only showcases the robust generalization of the models, but also reduces the need for extensive task-specific data, simplifying deployment in various real-world scenarios\cite{radford2021learning}.
The deployment of foundational models presents significant challenges. Their training requires substantial computational resources, which presents sustainability concerns. Moreover, using imbalanced datasets can perpetuate biases, leading to ethical issues. Additionally, the absence of standardized benchmarks makes it difficult to assess the effectiveness of these models in various tasks. Addressing these issues requires ongoing research to develop more efficient training methods, ensure fairness, and create reliable evaluation metrics\cite{awais2023foundational}.

\section{Method}

\subsection{Problem formulation}
Consider a series of \( T \) tasks denoted by \( \mathbf{Q} = \{\mathcal{Q}^1, \mathcal{Q}^2, \cdots, \mathcal{Q}^{T}\} \). Here, \( \mathcal{Q}^1 \) represents the base task, and the subsequent tasks are novel tasks that are incrementally added. $\mathcal{C}^{t}$ signifies the label space i.e, the classes within task $\mathcal{Q}^{t}$ during training. Note that the training label spaces between different tasks are disjoint, i.e., for any $i, j \in [1,T]$ and $i \neq j $,  $\mathcal{C}^{i}\cap \mathcal{C}^{j}= \emptyset$. Each task's classes are linked with prompt descriptions, noted as $\mathcal{P}^{t}$. Therefore, each task can be depicted as a tuple $\mathcal{Q}^t = \left \{\mathcal{X}^{t}_{i},\textbf{y}^{t}_{i},\textbf{p}^{t}_{i} \right \}_{i=1}^{n_{t}}$, where $\mathcal{X}^{t}_{i} = \{\textbf{x}_{i,j}^{t}\}_{j = 1}^{l}$ represents a 3D point cloud object with coordinates  $\textbf{x}_{i,j}^{t}\in\mathbb{R}^{3}$. Furthermore, $\textbf{y}^{t}_{i} \in \mathcal{C}^{t}$  and  $\textbf{p}^{t}_{i} \in \mathcal{P}^{t}$ denote the label of the point cloud and its associated class prompt description, respectively. Within the FSCIL framework, for the base task $\mathcal{Q}^1$, the model undergoes training on a large-scale synthetic 3D dataset. As for $t>1$, training data are sourced from real-world 3D point clouds with only a few instances. The model is trained sequentially across tasks $t = 1, \dots, T$. However, during the training of the $t$-th task $\mathcal{Q}^t$, the model encounters $\mathcal{X}^t$,  $\textbf{y}^t$, and $\{\mathcal{P}^1, \mathcal{P}^2, ..., \mathcal{P}^t\}$. During inference, the model trained on the current task $\mathcal{Q}^{t}$ is anticipated to classify test samples from both the current and preceding tasks, namely $\{\mathcal{Q}^1, \mathcal{Q}^2, \cdots, {\mathcal{Q}}^t\}$.

\subsection{Model overview}

Given the input sample \(\mathcal{X}_{i}^{t}\), its feature representation \(\textbf{v}^{t}_{i} \in \mathbb{R}^{m}\) is extracted using the vision encoder \(V_{e}\). The prompts \(\{\textbf{p}_{1}, \textbf{p}_{2}, \cdots, \textbf{p}_{C}\}\) of all classes from the current and preceding tasks are then processed through the text encoder \(T_{e}\), resulting in the feature representations \(\{\textbf{e}_{1}, \textbf{e}_{2}, \cdots, \textbf{e}_{C}\}\), with \(\textbf{e}_{j} \in \mathbb{R}^{m}\). Next, the vision and text features are concatenated and fed into an alignment module \( A \). This module connects features from two different modalities: vision and language. Specifically, the alignment module \(A\) generates a scalar value \({a}^{t}_{ij}\) ranging from 0 to 1, serving as a measure of the similarity between the visual and prompt feature embeddings. We then construct a similarity vector between the point cloud feature \(\textbf{v}^{t}_{i}\) and all class candidate features \(\{\textbf{e}_{1}, \textbf{e}_{2}, \cdots, \textbf{e}_{C}\}\) as follows: \(\textbf{a}^{t}_{i}= \{{a}^{t}_{i1}, {a}^{t}_{i2}, \cdots, {a}^{t}_{iC}\}\). This similarity vector \(\textbf{a}^{t}_{i}\) is then fed into an adaptor module comprising two caches: a base task cache \(B\) and a novel task cache \(N\). The base task cache contains test samples of the base task, selected based on a policy during inference to control forgetting of the base task. The novel task cache comprises few-shot training samples of novel classes to help the model learn new classes without fine-tuning. Finally, the updated similarity vector \(\textbf{b}^{t}_{i}\) is extracted from the adaptor module and merged with the original \(\textbf{a}^{t}_{i}\) to construct the final score \(\textbf{z}^{t}_{i}\).

\begin{figure*}[!t]
\centering
\begin{center} 
\includegraphics[width=1\textwidth]{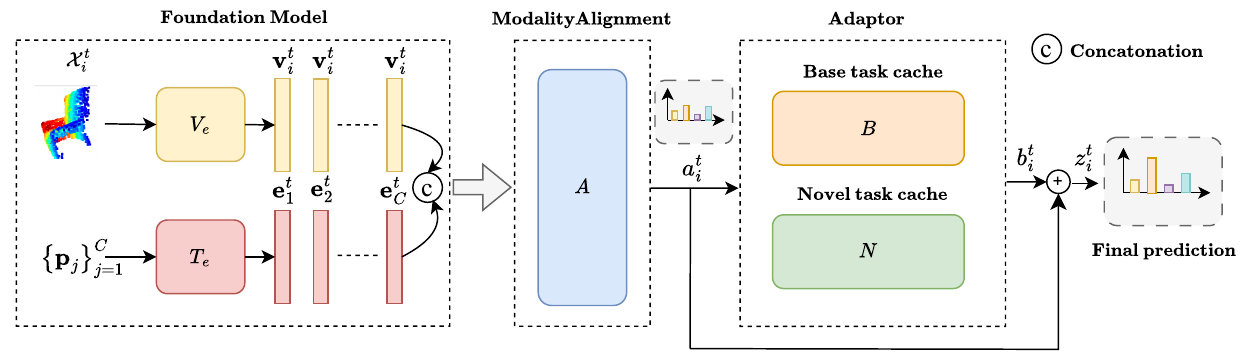} 
\end{center}
\caption{Feature \(\textbf{v}^{t}_{i} \in \mathbb{R}^{m}\) is extracted from input \(\mathcal{X}_{i}^{t}\) using vision encoder \(V_{e}\). Prompts \(\{\textbf{p}_{1}, \textbf{p}_{2}, \cdots, \textbf{p}_{C}\}\) are processed via text encoder \(T_{e}\) to obtain features \(\{\textbf{e}_{1}, \textbf{e}_{2}, \cdots, \textbf{e}_{C}\}\). These features are concatenated and aligned by module \(A\), producing similarity vector \(\textbf{a}^{t}_{i}\). This vector is refined by an adaptor module with base task cache \(B\) and novel task cache \(N\), resulting in the final score \(\textbf{z}^{t}_{i}\).}
\label{model_overview}
\end{figure*}

\subsection{Foundation model}
In recent years, vision-language foundation models have gained significant attention in computer vision tasks~\cite{zhang2022pointclip, zhu2023pointclip,radford2021learning}. This paper uses the Uni3D vision-language 3D foundation model~\cite{zhang2023uni3d}, exclusively trained on a substantial corpus of point cloud-text pairs, as our backbone. This model includes a vision encoder \( V_{e} \), responsible for extracting features from input point cloud data \( \mathcal{X}_{i}^{t} \), and a text encoder \( T_{e} \), which generates embeddings for the input class prompt description \( \textbf{p}_{i} \). The outputs of the vision encoder \( V_{e} \) and the text encoder \( T_{e} \) are denoted as \( \textbf{v}^{t}_{i} \in\mathbb{R}^{m} \) and \( \textbf{e}_{j} \in\mathbb{R}^{m} \), respectively. Furthermore, these outputs are aligned in the same embedding space. However, additional alignment between vision and text modalities is required for the FSCIL task, using training samples from the base task \( \mathcal{C}^{1} \).

\subsubsection{Modality alignment:}
To further align the vision and language branches for the downstream task in the FSCIL setting, we train an alignment module \(A\) using a training sample of the base task $\mathcal{C}^{1}$. This alignment module, also referred to as a relation module~\cite{8578229} in the literature, provides a similarity score between \([0, 1]\). For the alignment module, we use three fully connected layers with 2048, 1024, and 1 hidden units. For each training sample, we generate a score for each class of the base task as \(a_{ij}^t = \gamma \circ A \circ (\textbf{v}^t_{i} \oplus {\textbf{e}}_{j}), j \in \mathcal{C}^{t}\), $t=1$, where \(\oplus\) is the concatenation operator, \(A\) is the alignment module, and \(\gamma\) is the sigmoid function. For each feature \(a_{ij}\) and the corresponding ground truth \(\textbf{y}_i\), we train the \(A\) module for the base task using the binary cross-entropy cost function as follows:
\begin{align}
L_{r} = -\frac{1}{\left | \mathcal{S} \right |}\sum_{\textbf{y}_{i} \in \mathcal{S}}\bigg(\textbf{1}(y_{i}^{t} == k)\mathrm{log}(a_{ik}^{t}) + \big(1 - \textbf{1}(y_{i}^{t} == k)\big)\mathrm{log}(1 - a_{ik}^{t})\bigg),
\end{align}

\noindent where \(\mathcal{S}\) denotes the set of true labels in the base task. The trained alignment module \( A \) will be frozen for few-shot novel tasks.

\begin{figure*}[!t]
\centering
\begin{center}
\includegraphics[width=1\textwidth]{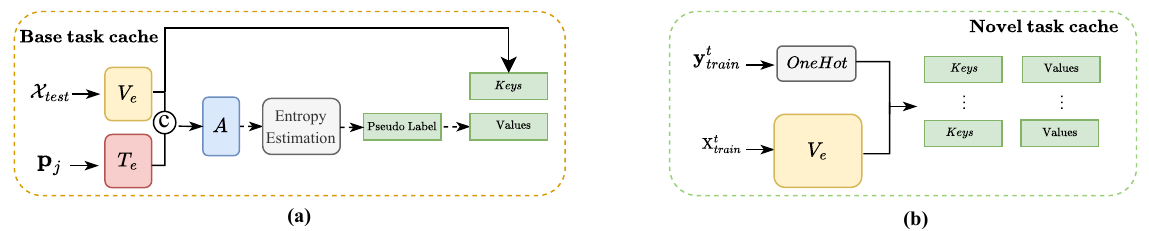}  
\end{center}
\caption{(a) Base task cache: This cache stores test samples from the base task to address forgetting issues, selecting samples based on their entropy values. The cache updates when a new test sample has a lower entropy than those currently stored. (b) Novel task cache: This cache contains training samples from few-shot novel classes.}
\label{Adaptor}
\end{figure*}

\subsection{Training-free adapator and Inference}

We propose a novel adaptor module to address the issue of forgetting base task classes while accommodating novel classes without encountering overfitting. This module comprises two caches containing key-value pairs representing the base and novel classes. Specifically, one cache stores features related to base task samples along with their pseudo-labels. At the same time, the other accumulates few-shot training samples from task 1 to the current task with their labels. During inference, test sample features act as queries to retrieve the most relevant information stored in the caches. Leveraging this retrieved information, the output for each test sample is adapted to optimize performance. Importantly, this caching approach does not require additional parameters or training.

\subsubsection{Base task cache:} The base task cache $B$ consists of key-value pairs organized as a dynamic queue for each class. It aims to store features from the base task test samples as keys that produce high-quality pseudo-labels. Initially, this cache is empty for each class and then filled with appropriate key-value pairs during the inference of the base task. Given the capacity of samples that can be stored for each class in the cache, this method gradually incorporates test predictions with lower entropy to maintain high-quality pseudo-labels. Consider the text and vision encoder networks denoted as \( T_{e} \) and \( V_{e} \), respectively. For all classes in the base task, we compute the text features using predefined prompts \( \textbf{p}_{j} \in \mathcal{P}^{1} \). Each test sample is also processed by the vision encoder to obtain representations \( v_{i} \). To construct the base cache, a pseudo-label, which is a one-hot vector from a categorical distribution, is generated for each test data \( \mathcal{X}_{i} \) by applying the softmax function on the output logits derived from the combination of text and vision features obtained through the alignment module $A$. This pseudo-label, along with its corresponding vision feature, must satisfy two conditions to be placed in the cache: 1) The capacity of the number of samples for that class \( \hat{L}_B \) has not been reached. In this case, the pseudo-label \( l \) along with the corresponding \( v_{i} \) is added to \( Q_{B} \) and \( L_{B} \) as a key-value pair for that class. 2) If the capacity has been reached, we check whether the new sample has a lower entropy than the existing samples in the cache. If it does, it replaces the sample with the highest entropy, \( \{ \textbf{q}^{\text{ent}}, \hat{\ell}^{\text{ent}}_b
 \} \).
\begin{equation}
H(\mathbf{a}^{\text{t}}_i) < H(\mathbf{a}^{\text{ent}}_i).
\end{equation}
Here, \( H \) denotes the entropy function, indicating the level of uncertainty. By considering these two conditions, in addition to adhering to the sample capacity for each class, we ensure that the pseudo-labels in the cache are of high quality.

\subsubsection{Novel task cache:} For the novel task cache \( N \), we employ feature embeddings extracted from few-shot training samples of newly introduced classes. With \( K \)-shot training samples available per class in the novel task, we aim to construct a key-value cache module as an adapter. Each sample undergoes feature extraction using the vision encoder. These extracted features from the vision encoder, paired with their respective class labels, are then integrated into \( Q_n \) and \( L_n \) as key-value pairs for each class.

\subsubsection{Inference:} During inference, the cache module, which includes key-value pairs obtained from the base cache and the novel cache, utilizes the features obtained from the vision encoder for the input test data sample as a query. It checks which of the stored features in the cache has the highest match with this query. In that case, it uses the information from the corresponding key-value pair to retrieve the results from the prediction output generated by the relation module for this data. The adaptive prediction vector \(\textbf{b}^{t}_{i}\) using the cache is obtained as follows:

\begin{equation}
P_{\text{cache}}(\textbf{v}_{i}^{t}) = A(\textbf{v}_{i}^{t} \mathbf{Q}^T) \mathbf{L},
\end{equation}

where \( A \) is the adaptation function introduced by \cite{zhang2022tip}:

\begin{equation}
A(u) = \exp(-\beta(1 - u)), \quad \text{where} \quad u \in [0, 1].
\label{eq:affinity}
\end{equation}
and the one-hot vector \(\mathbf{L}\) represents the stored value for each key, derived from the corresponding cache information.
Thus, the output \(\textbf{a}_{i}^{t}\) obtained for the test sample is updated by the cache, and the final output is computed as follows:

\begin{equation}
\textbf{z}_{i}^{t} = \textbf{a}_{i}^{t} + \alpha \textbf{b}_{i}^{t}.
\label{eq:final_prediction}
\end{equation}

\section{Experiments}

\noindent\textbf{Datasets.} Our paper leverages four distinct 3D datasets, which include synthetic structures (ModelNet \cite{wu20153d} and ShapeNet \cite{chang2015shapenet}) as well as real-scanned datasets (ScanObjectNN \cite{uy2019revisiting} and CO3D \cite{reizenstein2021common}). Adhering to the experimental setup proposed by \cite{chowdhury2022few}, our framework is designed to facilitate cross-dataset incremental learning with focused classifications. These experiments aim to bridge the gap between synthetic and scanned data by establishing base classes in synthetic datasets and gradually introducing classes from scan-derived datasets. Detailed experimental configurations are provided in Table~\ref{tab:experiment_setups}.

\begin{table}[!t]
\centering
\caption{Summary of our experimental setups.}
\label{tab:experiment_setups}
\begin{tabular}{@{}lcccccc@{}} 
\toprule
Experiment Setups & \# Base & \# Novel & \# Tasks & \# Train & \# Test & \# Test \\
                  & Classes & Classes  &          & in Base  & in Base & in Novel \\
\midrule
ModelNet40 $\rightarrow$ ScanObjectNN & 26  & 11  & 4   & 4999  & 1496 & 475 \\
ShapeNet $\rightarrow$ ScanObjectNN   & 44  & 15  & 4   & 22797 & 5845 & 581 \\
ShapeNet $\rightarrow$ CO3D           & 39  & 50  & 11  & 26287 & 6604 & 1732 \\
\bottomrule
\end{tabular}
\end{table}

\noindent\textbf{Implementation details.} In our implementation, we employ the pre-trained text encoder component of the `EVA02-E-14-plus' CLIP model\cite{sun2023eva} to extract feature embedding from class names within our dataset. For processing point cloud data, we utilize the `base' scale configuration of the Uni3D architecture~\cite{zhou2023uni3d}, specifically the `eva02\_base\_patch14\_448' model as our point cloud encoder. This model choice aligns with the scalability principles outlined by \cite{zhang2023uni3d}, effectively balancing computational efficiency with the ability to capture detailed spatial features. Equipped with 88 million parameters, the `Base' model optimizes our computational resources while ensuring comprehensive 3D data representation. The point cloud and text encoders are initialized with pre-trained weights, which are frozen during training. Furthermore, we incorporate a trainable alignment module as defined by \cite{8578229} to integrate the point cloud and text features. This alignment network comprises three fully connected layers with configurations of 2048, 1024, and 1 neurons, respectively. LeakyReLU activations are used in the initial layers, while the output layer employs a Sigmoid activation. The alignment module is specifically trained as a feature extractor for task 0, utilizing basic data across 10 epochs. We employ the Adam optimizer, setting a learning rate of 0.001 and a batch size 25. Additionally, we maintain a cache of five key samples and their corresponding values for each task, incrementally building this dataset. Our experiments use the PyTorch framework on a single NVIDIA A100 GPU.

\noindent\textbf{Evaluation metrics.} In each incremental phase, we assess the accuracy by considering both base and novel classes. Following the approach outlined in \cite{graph-few-shot2022}, we then determine the rate of accuracy decline, denoted as $\Delta = \frac{|acc_{T}-acc_{0}|}{acc_{0}} \times 100$. Here, $acc_{T}$ is the accuracy at the final task, while $acc_{0}$ is the accuracy at the outset. The parameter $\Delta$ provides a consolidated measure of the method's efficacy, with a lower $\Delta$ indicating superior performance. This evaluation is based on the average accuracy calculated over ten trials, each with a different random seed. The accuracy and $\Delta$ metrics cannot precisely evaluate the balance between forgetting old class samples and learning novel classes. This is because a large portion of the dataset consists of base classes, and only a small number of training samples are used for new classes. Therefore, even if the model does not perform well on new classes but achieves good accuracy for the base task, it can still report good numbers for both metrics. To better assess how well our model retains knowledge of base tasks and performs on new classes, we use the metric introduced in paper~\cite{peng2022few}, known as harmonic accuracy. This metric is calculated based on the following formula: $A_h = \frac{2 \times A_b \times A_n}{A_b + A_n}$, where $A_b$ is the accuracy of the base classes and $A_n$ stands for the accuracy of new classes. Additionally, we report the performance of the base classes and the new classes in each learning session. The higher the harmonic accuracy, the better the network maintains a balance between the accuracy of old and novel classes.

\begin{table*}[!t]
\caption{\small Summary of FSCIL results.}
\newcolumntype{C}[1]{>{\centering\arraybackslash}p{#1}}
\centering \small

\scalebox{0.6}{
\begin{minipage}{.88\textwidth}
\begin{tabular}{C{2.2cm}|C{0.6cm}C{0.6cm}C{0.6cm}C{0.6cm}C{0.6cm}C{0.6cm}C{0.6cm}C{0.6cm}C{0.6cm}C{0.6cm}C{0.6cm}C{0.6cm}}
\hline
& \multicolumn{12}{c}{ShapeNet $\rightarrow$ CO3D}\\
\hline
Method & 39 & 44 & 49 & 54 & 59 & 64 & 69 & 74 & 79 & 84 & 89 & $\Delta\downarrow$\\
\hline
\textit{FT} & 81.0 & 20.2 & 2.3 & 1.7 & 0.8 & 1.0 & 1.0 & 1.3 & 0.9 & 0.5 & 1.6 & 98.0\\
\textit{Joint} & 81.0 & 79.5 & 78.3 & 75.2 & 75.1 & 74.8 & 72.3 & 71.3 & 70.0 & 68.8 & 67.3 & 16.9\\
\hline
LwF \cite{li2017learning} & 81.0 & 57.4 & 19.3 & 2.3 & 1.0 & 0.9 & 0.8 & 1.3 & 1.1 & 0.8 & 1.9 & 97.7\\
IL2M \cite{belouadah2019il2m} & 81.0 & 45.6 & 36.8 & 35.1 & 31.8 & 33.3 & 34.0 & 31.5 & 30.6 & 32.3 & 30.0 & 63.0\\
ScaIL \cite{belouadah2020scail} & 81.0 & 50.1 & 45.7 & 39.1 & 39.0 & 37.9 & 38.0 & 36.0 & 33.7 & 33.0 & 35.2 & 56.5\\
EEIL \cite{castro2018end} & 81.0 & 75.2 & 69.3 & 63.2 & 60.5 & 57.9 & 53.0 & 51.9 & 51.3 & 47.8 & 47.6 & 41.2\\
FACT \cite{zhou2022forward} & 81.4 & 76.0 & 70.3 & 68.1 & 65.8 & 63.5 & 63.0 & 60.1 & 58.2 & 57.5 & 55.9 & 31.3\\
Sem-aware \cite{cheraghian2021semantic} & 80.6 & 69.5 & 66.5 & 62.9 & 63.2 & 63.0 & 61.2 & 58.3 & 58.1 & 57.2 & 55.2 & 31.6\\
Microshape~\cite{chowdhury2022few}  & {82.6} & {77.9} & {73.9} & {72.7} & {67.7} & {66.2} & {65.4} & {63.4} & {60.6} & {58.1} & {57.1} & {30.9} \\
C3PR \cite{cheraghiancanonical} & 83.6 & 80.0 & 77.8 & 75.4 & 72.8 & 72.3 & 70.3 & 67.9 & 64.9 & 64.1 & 63.2 & 24.4  \\
Ours & \textbf{87.3} & \textbf{86.2} & \textbf{84.4} & \textbf{82.2} & \textbf{80.7} & \textbf{79.6} & \textbf{78.2} & \textbf{76.8} & \textbf{76.1} & \textbf{74.5} & \textbf{72.6} & \textbf{16.8} \\
\hline
\end{tabular}
\end{minipage} \hfill 
\begin{minipage}{.34\textwidth}
\centering \small
\begin{tabular}{C{0.65cm}C{0.65cm}C{0.65cm}C{0.65cm}C{0.65cm}}
\hline
\multicolumn{5}{c}{ ModelNet $\rightarrow$ ScanObjectNN }\\\hline
26 & 30 & 34 & 37 & $\Delta\downarrow$\\
\hline
88.4 & 6.4 & 6.0 & 1.9 & 97.9\\
88.4 & 79.7 & 74.0 & 71.2 & 19.5\\
\hline
88.4 & 35.8 & 5.8 & 2.5 & 97.2\\
88.4 & 58.2 & 52.9 & 52.0 & 41.2\\
88.4 & 56.5 & 55.9 & 52.9 & 40.2\\
88.4 & 70.2 & 61.0 & 56.8 & 35.7\\
89.1 & 72.5 & 68.3 & 63.5 & 28.7\\
88.5 & {73.9} & 67.7 & 64.2 & 27.5\\
\textbf{89.3} & 73.2 & {68.4} & {65.1} & {27.1}\\
88.3 & 75.7 & 70.6 & 67.8 & 23.2\\
87.7  & \textbf{84.7} & \textbf{81.5} & \textbf{79.2} & \textbf{9.6}\\

\hline
\end{tabular}
\end{minipage} \hfill 
\begin{minipage}{.3\textwidth}
\begin{tabular}{C{0.65cm}C{0.65cm}C{0.65cm}C{0.65cm}C{0.65cm}}
\hline
\multicolumn{5}{c}{ShapeNet $\rightarrow$ ScanObjectNN}\\\hline
44 & 49 & 54 & 59  & $\Delta\downarrow$\\
\hline
81.4 & 38.7 & 4.0 & 0.9 & 98.9\\
81.4 & 82.5 & 79.8 & 78.7 & 3.3\\
\hline
81.4 & 47.9 & 14.0 & 5.9 & 92.8\\ 
81.4 & 53.2 & 43.9 & 45.8 & 43.7 \\
81.4 & 49.0 & 46.7 & 40.0 & 50.9 \\
81.4 & 74.5 & 69.8 & 63.4 & 22.1\\
82.3 & 74.6 & 69.9 & 66.8 & 18.8\\
81.3 & 70.6 & 65.2 & 62.9 & 22.6\\
{82.5} & {74.8} & {71.2} & {67.1} & {18.7}\\
84.5 & 77.8 & 75.5 & 71.9 & 14.9\\
\textbf{90.8} & \textbf{86.5} & \textbf{86.4} & \textbf{85.6} & \textbf{5.75}
\\
\hline
\end{tabular}
\end{minipage}
} 

\label{table:cross}
\end{table*}

\subsection{Main results}

%%%%% CROSS DATASET %%%%%

In this section, we compare our method against several state-of-the-art(SOTA) approaches, including FT, Joint, LwF \cite{li2017learning}, IL2M \cite{belouadah2019il2m}, ScaIL \cite{belouadah2020scail}, EEIL \cite{castro2018end}, FACT \cite{zhou2022forward}, Sem-aware \cite{cheraghian2021semantic}, Microshape \cite{chowdhury2022few}, cross-domain~\cite{Tan_2024_WACV} and C3PR \cite{cheraghiancanonical}. FT (Fine-Tuning) involves fine-tuning the model on new classes without revisiting old classes, often leading to catastrophic forgetting. Jointly retraining the model on all classes, assuming access to all data is often impractical. State-of-the-art methods such as IL2M~\cite{belouadah2019il2m}, ScaIL~\cite{belouadah2020scail}, EEIL~\cite{castro2018end}, LwF~\cite{li2017learning}, FACT~\cite{zhou2022forward}, and Sem-aware~\cite{cheraghian2021semantic} were initially reported on 2D datasets. We adapted their implementations using PointNet features for 3D datasets. Our results are summarized in Table \ref{table:cross}. Our observations are as follows:
Due to the presence of noise in the 3D real-can dataset, achieving satisfactory performance poses significant challenges. FT shows the lowest accuracy across all datasets, with a $\Delta$ of 98.0 for ShapeNet to CO3D and 97.2 for ModelNet to ScanObjectNN. This significant drop is attributed to catastrophic forgetting, as the model is fine-tuned on new classes without revisiting old ones. Conversely, Microshape \cite{chowdhury2022few} achieves the highest accuracy due to its innovative use of Microshape descriptions and their alignment with semantic prototypes, effectively minimizing domain gaps and providing superior results in each incremental task. IL2M \cite{belouadah2019il2m} and ScaIL \cite{belouadah2020scail} propose special training mechanisms tailored for 2D image examples. However, their performance drops when applied to 3D datasets due to the inherent complexities and noise in 3D data. Both LwF \cite{li2017learning} and EEIL \cite{castro2018end} apply knowledge distillation in their loss functions, with EEIL \cite{castro2018end} enhancing LwF \cite{li2017learning} by additionally using exemplars. Despite these enhancements, they struggle with the challenges posed by 3D data, leading to higher performance degradation. FACT addresses few-shot class incremental learning through feature augmentation and classification tuning, while Sem-aware \cite{cheraghian2021semantic} successfully incorporates class-semantic embedding information during training, providing a marginal boost. However, both methods still fall short in 3D scenarios compared to our approach. Overall, these state-of-the-art methods primarily target 2D image data and fail to address the specific challenges of 3D data, such as noise and the need for robust spatial feature extraction. C3PR \cite{cheraghiancanonical} uses a combination of learned projections, model reprogramming, and prompt engineering to tackle FSCIL for 3D point cloud objects. Overall, our approach significantly outperforms other methods in terms of accuracy.

\begin{figure}[!t]
\centering
\includegraphics[width=1\linewidth]{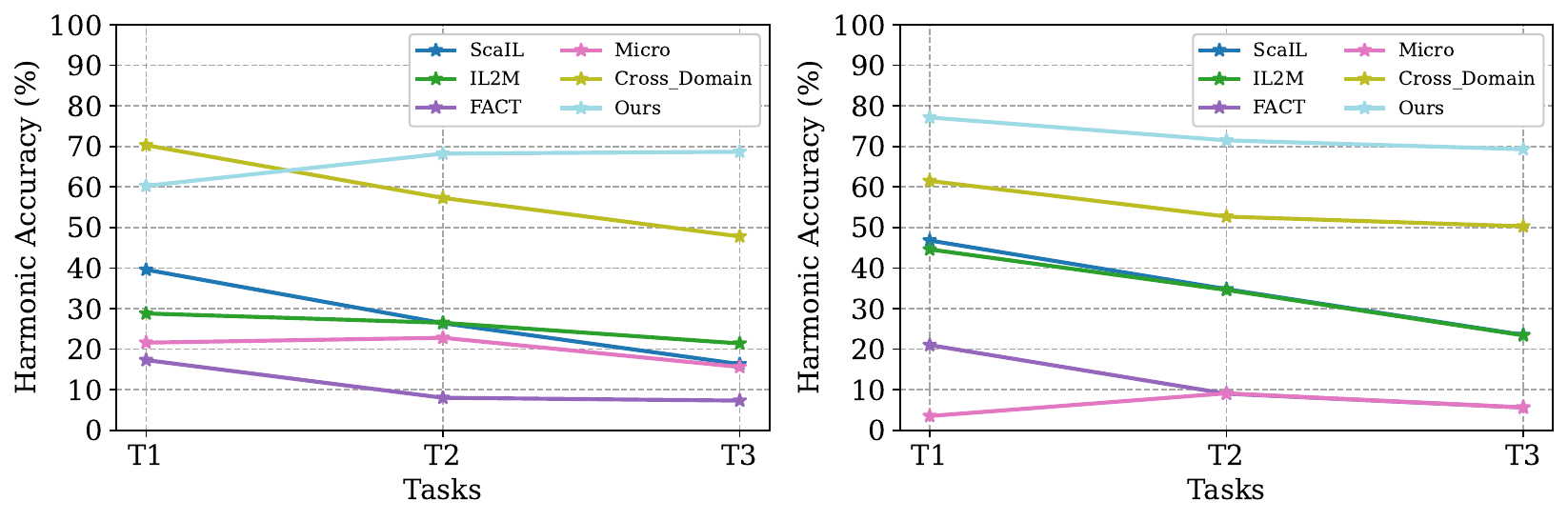}
\caption{ Comparison of the harmonic accuracy with SOTA methods on ShapeNet to ScanObjectNN and ModelNet40 to ScanObjectNN datasets.}
\label{fig:compare harmonic accuracy}
\end{figure}

In FSCIL, achieving high performance on both base and novel classes is essential. To assess this, we compare our proposed method with state-of-the-art approaches using the harmonic mean metric. Higher values in this metric indicate effective performance across both base and novel test samples, while a decrease suggests poorer performance on either base or novel tasks. It is worth noting that this evaluation method was introduced by \cite{Tan_2024_WACV} and is referred to as the cross-domain method. As shown in Fig.~\ref{fig:compare harmonic accuracy}, our proposed method significantly outperforms all other methods on both ShapeNet-to-ScanObjectNN and ModelNet-to-ScanObjectNN datasets.

\subsection{Ablation study}

In this section, we conduct ablation studies to evaluate the effectiveness of our designs. All ablation studies are performed on the ShapeNet to ScanObjectNN dataset, where our method achieves an accuracy of 85.6\% for the final task under the default settings.

\begin{figure}[!t]
\centering
\includegraphics[width=1\linewidth]{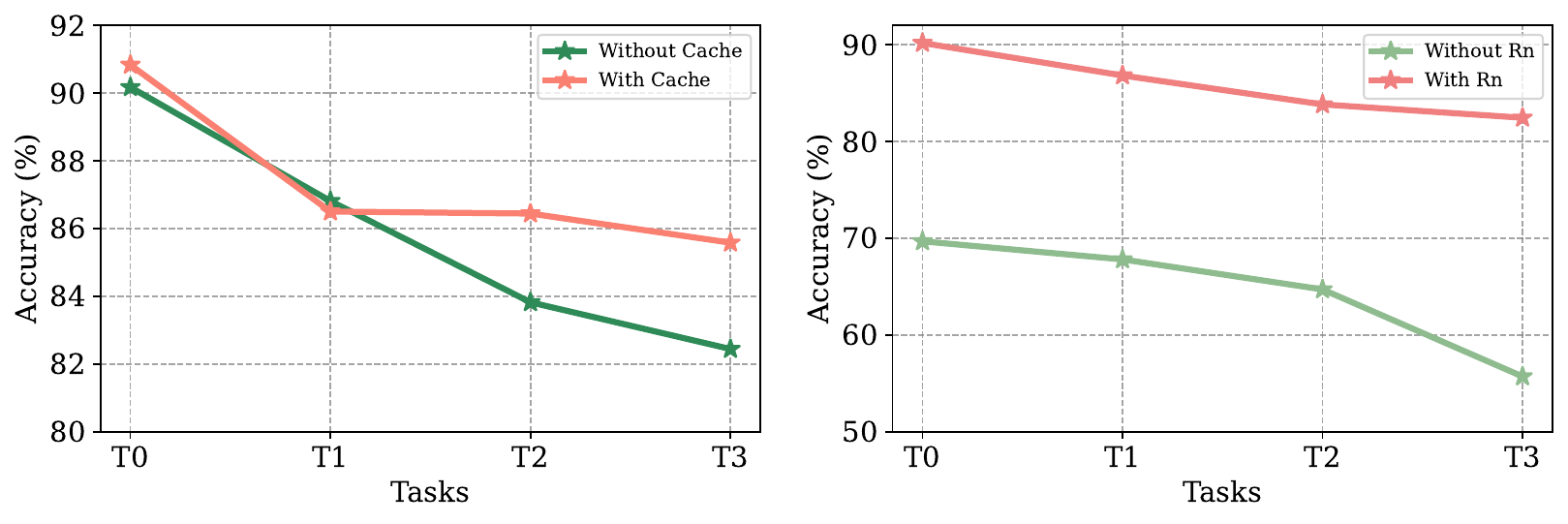}
\caption{ The influence of caches(a) and the impact use relation module after encoders vs zeroshot(b).}
\label{fig:impact cache}
\end{figure}

\noindent\textbf{The impact of cache:} In Fig. \ref{fig:impact cache}(a), we observe a comparison of the model's performance with and without the use of the cache. When our model relies solely on the predictions obtained from the alignment module $A$, the model's accuracy in predicting the base task data does not suffer from the forgetting problem as the number of tasks increases. However, the model's performance on novel classes significantly drops. For tasks 1 to 3, the values of $A_n$ are 10.08, 5.9, and 4.6, respectively, indicating that the model is incapable of predicting new tasks and $A_{cct}$ is a result of the model's good performance on the base task due to training the relation module with the data from this task. However, when we use caches in the adaptor module to adapt the model's predictions, we observe that for incremental tasks 1 to 3, the values of $A_n$ are 45.8, 55.4, and 56.1, respectively. Consequently, we achieve an average harmonic accuracy of 67.73\%, indicating a balance between predicting new class data and not forgetting the previous task data.

\noindent\textbf{The role of alignment module:} In Fig. \ref{fig:impact cache}(b), we present the results obtained with and without using the alignment module. Suppose that we do not use the Relation Module to classify the results obtained from both encoders. In that case, we are in a zero-shot learning scenario since there are no parameters to train, and we only use the point cloud encoder and text encoder with pre-trained and frozen weights. In this case, the output for each sample is obtained by calculating the maximum cosine similarity between the outputs of the text encoder and the point cloud encoder. The alignment module is also evaluated when trained only for the base task, and no samples are stored in the cache. The use of the Relation Module allows for combining the features obtained from both encoders, resulting in a better-learned feature space.

\begin{figure}[!t]
\centering
\includegraphics[width=1\linewidth]{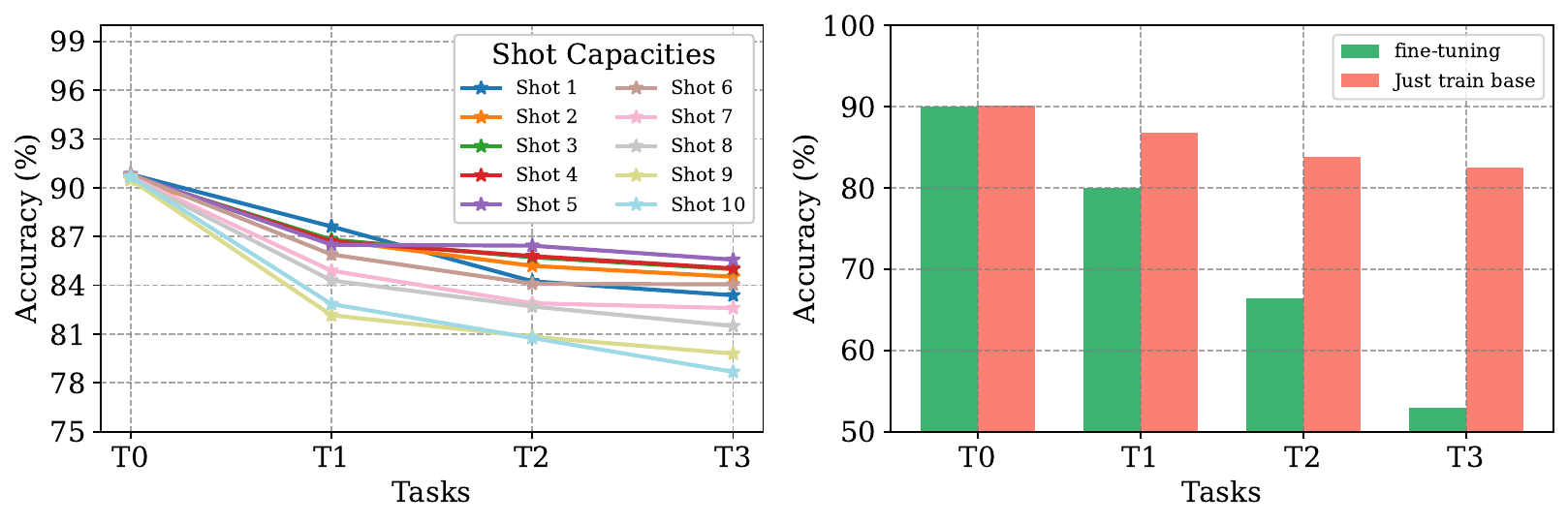}
\caption{ (a) The influence of the number of samples in the cache. (b) The effect of finetuning the relation module in each task versus training solely for the base task and then freezing.}
\label{fig:impact shotcapacity}
\end{figure}

\noindent\textbf{The impact of the number of samples in the cache:} We studied the effect of the shot capacity, which refers to the maximum number of pairs of keys-values per class, both in the basic and the novel caches. The aim is to find the optimal balance between the diversity and accuracy of the key-value pairs. Considering 5 shots per class from the training data for each class except task zero, we examined cache construction from size 1 to 10. As explained in the previous section, the selection of each sample for the test cache is based on entropy, while the training cache is selected randomly. Given that random selection might affect the results, we repeated our experiments three times in this section and reported the average results. The results shown in Fig.~\ref{fig:impact shotcapacity}(a) indicate that increasing the cache size improves accuracy until the entire training data fits into the cache. However, accuracy decreases when the number of test cache data exceeds the training data. This decrease in accuracy results from the base cache data, including pseudo-labels obtained based on the lowest entropy of the model's predictions. Consequently, these data are accompanied by noise.

\noindent\textbf{Full fine-tuning vs only base task training:} In Fig. \ref{fig:impact shotcapacity}(b), we examine the effect of fine-tuning the alignment module for novel classes. Making the alignment module trainable for all tasks leads to a decrease in accuracy. This is because training the network with few-shot data during incremental stages enhances overfitting, causing the network weights to shift towards learning the classes of the new tasks, thus forgetting the previous tasks. However, if the alignment module is only trained for the base task and then frozen for subsequent tasks, we observe that the accuracy for the base task is maintained.

\noindent\textbf{The impact of {$\alpha$}  and {$\beta$}:} The hyperparameter $\alpha$ used in Equation \ref{eq:affinity}  controls the extent to which new predictions obtained from the cache module are combined with predictions from the relation module's output. A larger $\alpha$ indicates greater importance given to the knowledge obtained from the cache data. To select an appropriate $\alpha$, we evaluated the performance of the model based on the mean harmonic accuracy between tasks, which is a more precise metric than the plain accuracy. With $\beta$ set to 2, we varied $\alpha$ from 0 to 3. An $\alpha$ of zero means that the cache module's knowledge is not used at all, effectively resulting in a model without a cache. Next, we examine the hyperparameter $\beta$ used in Equation \ref{eq:final_prediction}. This parameter controls the sharpness of similarity. When $\beta$ is large, only the most similar training samples to the test image in the embedding space significantly affect the prediction and vice versa. With $\alpha$ set to 2, we varied $\beta$ from 0 to 3. Table \ref{table:alpha_beta} shows that the optimal mean harmonic accuracy is achieved when both $\alpha$ and $\beta$ are set to 2. Therefore, the knowledge obtained from cache data significantly contributes to achieving desirable results in multi-class incremental learning without the need for additional training.

\begin{table}[!t]
\centering
\caption{\small Ablation studies of impact {$\alpha$} and {$\beta$} on mean Harmonic Accuracy.}
\begin{tabular}{llllccccc|llllccccc}
\hline
\multicolumn{4}{l}{Residual Ratio $\alpha$} & 0 & 0.5 & 1 & \textbf{2} & 3 & \multicolumn{4}{l}{Sharpness Ratio $\beta$} & 0 & 0.5 & 1 & \textbf{2} & 3 \\
\multicolumn{4}{c}{HM} & 12.6 & 35.3 & 53.2 & \textbf{67.7} & 65.0 & \multicolumn{4}{c}{HM} & 6.3 & 58.2 & 64.5 & \textbf{67.7} & 65.2 \\ \hline
\end{tabular}
\label{table:alpha_beta}
\end{table}

\section{Discussion} 

\noindent\textbf{The impact of foundation model:} In our approach, we harness the capabilities of a 3D vision-language foundation model~\cite{zhou2023uni3d}, which significantly enhances the performance of our method. This observation underscores the broader applicability of 3D foundation models to tackle related downstream tasks under low data conditions, such as zero-shot learning, few-shot learning, and dealing with long-tailed distributions. These models demonstrate their utility by effectively leveraging semantic and structural information embedded in 3D data, thereby improving adaptability and generalization across diverse and challenging learning scenarios. This highlights their potential to advance various applications in 3D computer vision and beyond.

\noindent\textbf{Limitation:} Although our method has demonstrated state-of-the-art results in the 3D point cloud domain, it also highlights limitations that warrant discussion. Specifically, we have not fully capitalized on the potential of vision-language foundation models. Future research directions include exploring advanced fine-tuning techniques like prompt tuning strategies \cite{zhou2022cocoop, zhou2022coop}, LORA \cite{hu2022lora}, and enhanced prompt engineering using large language models (LLMs) such as GPT \cite{brown2020language} or in-context learning approaches \cite{NEURIPS2023_398ae57e}.

\section{Conclusion}
In conclusion, this paper presents a pioneering approach tailored to address the challenges of Few-Shot Continual Incremental Learning (FSCIL) in 3D computer vision. By leveraging a robust 3D foundation model trained on extensive point cloud data, we design a novel training-free adaptation module to effectively manage forgetting and overfitting issues inherent in FSCIL scenarios. Our method utilizes a dual cache strategy that optimally utilizes previous task test samples based on model confidence scores to maintain performance on base classes while integrating few-shot samples from new tasks to enhance generalization and prevent overfitting. The experimental results across diverse datasets, including ModelNet, ShapeNet, ScanObjectNN, and CO3D, demonstrate our approach's superior efficacy and versatility compared to existing FSCIL methods. This work contributes significantly to advancing the capabilities of 3D vision-language models in handling continual learning tasks, paving the way for more robust and adaptable solutions in real-world applications of 3D computer vision.

\noindent\textbf{Acknowledgement.} This work was supported by the North South University (NSU) Conference Travel and Research Grants (CTRG) 2023–2024 (Grant ID: CTRG-23-SEPS-20).

\bibliographystyle{splncs04}

\bibliography{main}
\end{document}